\documentclass[3p,times,procedia]{elsarticle}
\usepackage{ecrc}
\usepackage[bookmarks=false]{hyperref}
    \hypersetup{colorlinks,
      linkcolor=blue,
      citecolor=blue,
      urlcolor=blue}
\usepackage{subcaption} % Required for subfigure\usepackage{subcaption} % Required for subfigure

\usepackage{titlesec}

\titlespacing{\section}{0pt}{3ex plus 0.5ex minus 0.2ex}{1ex plus 0.2ex}
\titlespacing{\subsection}{0pt}{3ex plus 0.3ex minus 0.2ex}{0.8ex plus 0.2ex}

\volume{00}

\firstpage{1}

\journalname{Procedia Computer Science}

\runauth{W. Christian et al.}

\jid{procs}

\usepackage{amssymb}
\usepackage[figuresright]{rotating}
\begin{document}
\begin{frontmatter}



\dochead{The 10th International Conference on Computer Science and Computational Intelligence 2025}%%%
%% e.g. \dochead{17th International Conference on Dynamical Processes in Excited States of Solids}

\title{Leveraging IndoBERT and DistilBERT for Indonesian Emotion Classification in E-Commerce Reviews}


\author[]{William Christian} 
\author[]{Daniel Adamlu\corref{cor1}}
\author[]{Adrian Yu}
\author[]{Derwin Suhartono}

\address{Computer Science Department, School of Computer Science, Bina Nusantara University, Jakarta, Indonesia 11480}

\begin{abstract}
%%Text of abstract
% Emotion classification in the Indonesian language is a crucial task in Natural Language Processing (NLP), especially for e-commerce, where understanding customer sentiment can improve decision-making. This study aims to enhance Indonesian emotion classification performance by leveraging IndoBERT and DistilBERT, applying data augmentation, and implementing an ensemble method using bagging. We fine-tuned IndoBERT and DistilBERT, two transformer-based models trained for the Indonesian language, on the PRDECT-ID dataset, performed data augmentation techniques such as back translation and synonym replacement, and combined multiple IndoBERT models using bagging to improve performance. IndoBERT outperformed DistilBERT, achieving 80\% accuracy on Indonesian emotion classification after hyperparameter tuning, while the best bagging ensemble of IndoBERT models reached 79.77\%, showing only a marginal improvement. IndoBERT proves to be the most effective model for Indonesian emotion classification, with data augmentation significantly improving performance, but bagging does not provide substantial additional benefits; future research should explore alternative architectures and strategies to enhance generalization for Indonesian NLP tasks.

Understanding emotions in the Indonesian language is essential for improving customer experiences in e-commerce. This study focuses on enhancing the accuracy of emotion classification in Indonesian by leveraging advanced language models, IndoBERT and DistilBERT. A key component of our approach was data processing, specifically data augmentation, which included techniques such as back-translation and synonym replacement. These methods played a significant role in boosting the model's performance. After hyperparameter tuning, IndoBERT achieved an accuracy of 80\%, demonstrating the impact of careful data processing. While combining multiple IndoBERT models led to a slight improvement, it did not significantly enhance performance. Our findings indicate that IndoBERT was the most effective model for emotion classification in Indonesian, with data augmentation proving to be a vital factor in achieving high accuracy. Future research should focus on exploring alternative architectures and strategies to improve generalization for Indonesian NLP tasks.

\end{abstract}

\cortext[cor1]{Corresponding author. Tel.: +62-21-534-5830.}
\end{frontmatter}

%\correspondingauthor[*]{Corresponding author. Tel.: +0-000-000-0000 ; fax: +0-000-000-0000.}
\email{daniel.adamlu@binus.ac.id}

% \linenumbers
\vspace*{6pt}

\raggedbottom

%\enlargethispage{-7mm}
\section{Introduction}

Emotion is an essential and important part of how humans communicate with each other and understand more about the meaning of their message. As a result, emotion classification has become an important area of research in Natural Language Processing (NLP) \cite{b1}, with applications in fields such as sarcastic analysis \cite{b2}, health care \cite{b3}, consumer analysis \cite{b4}, and online learning \cite{b5}. Various Machine Learning models have been developed to tackle this task, and Deep Learning has contributed significantly to advancements in the field \cite{b6}.

Emotion classification is the process of identifying emotions such as happy, anger, sadness, or love expressed in text. It goes beyond basic sentiment analysis by capturing more detailed emotional meaning. For example, knowing whether a customer is angry or sad can help companies respond more appropriately.

As emotion classification continues to evolve, its role in addressing real-world challenges has become increasingly significant. Emotions play a central role in how humans express needs, concerns, and intentions especially in text-based communication. By enabling a deeper understanding of human emotions, it improves decision making, improves interactions, and fosters more effective problem solving across various domains. These advancements, summarized in Table 1, highlight the growing impact of emotion-aware systems in diverse real-world scenarios.

\raggedbottom

\begin{table}[htbp]
\caption{Real-World Applications of Emotion Classification}
\centering
\begin{tabular}{|p{4cm}|p{10cm}|}
\hline
\textbf{Area} & \textbf{Description} \\
\hline
Healthcare & Enhances early detection of mental health conditions, refines patient feedback analysis, and enables personalized treatment strategies. \cite{b38}\\
\hline
Consumer Analysis & Analyzes customer emotions to refine products, enhance user experiences, and optimize marketing strategies. \cite{b39}\\
\hline
Online Learning & Assesses student engagement, detects learning difficulties, and enables personalized learning support. \cite{b40}\\
\hline
\end{tabular}
\label{tab:real_world_applications}
\end{table}

Despite the rise of deep learning models, emotion classification still faces several challenges and restrictions. One of the main challenges of emotion classification is the shortage of high-quality datasets that are needed for developing a high-performance system \cite{b7}. These datasets are often unbalanced or have limited resources, making it difficult to train models effectively.

Another main challenge stems from the complexity of human emotion, which results in fuzzy emotion boundaries; for example, there is no clear line that divides the emotions of joy and love. Human emotions also tend to be more introverted, which may lead to incomplete emotional information; therefore, some researchers have explored emotion classification using multimodal information \cite{b7}.

In the context of the Indonesian language emotion classification, a dataset collected from the product reviews of an Indonesian e-commerce platform, has been developed for emotion and sentiment analysis \cite{b33}. This dataset provides valuable real-world data for improving emotion classification models.

The emergence and rapid development of online marketplaces have stimulated e-commerce transactions in Indonesia, with 73.7\% of the population having internet access, and e-commerce is among the reasons for rising internet usage \cite{b8}. Product reviews play a crucial role in influencing customer purchasing decisions. By incorporating emotion classification into product analysis, businesses can gain deeper insights into consumer sentiment, allowing vendors to identify areas for improvement and enhance their products accordingly \cite{b33}. Research conducted on the PRDECT-ID dataset has achieved a maximum accuracy of only 75\% for emotion classification, highlighting significant room for improvement in handling Indonesian text \cite{b35} \cite{b36}. This paper aims to explore methods to enhance the performance of IndoBERT and DistilBERT models for emotion classification in the Indonesian language.

This paper is structured as follows. First, we introduce the goals and challenges of emotion classification, followed by the importance of product reviews for business success. In section 2, this paper talks about the studies that have been relevant for doing emotion classification and comparisons between models in both English and Indonesian language. Section 3 presents the methodology used in this study, including dataset selection, data preprocessing, and the models implemented for emotion classification. Section 4 discusses the experimental setup, evaluation metrics, and results obtained from various models. Finally, Section 5 provides conclusions, key findings, and potential directions for future research.

\section{Related Works}

\subsection{Methods}

Understanding human emotions is essential for achieving accurate emotion classification. Shaver et al. stated that emotions are hierarchically structured, with six basic emotions—love, joy, anger, sadness, fear, and tentatively surprise—serving as the fundamental categories that are derived from more complex emotional states \cite{b9}.

Emotion classification has evolved significantly over the years. Earlier works for inferring word relations and adjectives were focused on creating dictionaries and rules to automatically categorize words \cite{b10} or texts \cite{b11}; these methods were preferred because of their simplicity and interpretability. Later, dictionary and rule-based methods were used for emotion classification, focusing on using the relation and association between words and sentences \cite{b12} \cite{b13}.

As feature extraction techniques became more relevant, traditional machine learning algorithms emerged as viable solutions for emotion classification in text-based tasks \cite{b14}. Several studies have explored the effectiveness of these methods. For instance, \cite{b15} utilized a Support Vector Machine (SVM) for sentiment analysis at both the message and expression levels on Twitter data. Similarly, \cite{b16} applied a Naïve Bayes classifier for tweet classification and further experimented with other traditional classification algorithms to enhance performance in emotion detection tasks.

In recent years, Deep Learning has been a hot topic for research and study in every aspect of Machine Learning for emotion classification methods such as Deep Neural Networks (DNN) and Convolutional Neural Network (CNN) obtains better results when being compared to Machine Learning Methods \cite{b17}. Recurrent Neural Network (RNN) allows classification that uses context from the previous layer \cite{b18} utilizes CNN for feature extraction and RNN for emotion classification. Unfortunately, RNN networks usually have a problem with vanishing gradient where when doing backpropagation, earlier neurons don't get updated much \cite{b19}. Hence, methods such as Long Short Term Memory (LSTM) \cite{b20} and GRU (Gated Recurrent Units) \cite{b21} are used for emotion classification.

Despite the advancement of LSTM and GRU, these models face a big challenge: the limited ability to model long-range dependencies \cite{b22}. To address this issue, Transformers \cite{b23} were introduced, revolutionizing the field of NLP by enabling parallelized training and utilizing self-attention mechanisms. As a result, transformer-based models are made, for example, BERT \cite{b24}, RoBERTa \cite{b25}, and DistilBERT \cite{b26}, which have achieved incredible performance for text-based Emotion Classification tasks. A BERT-based model is also trained and made to accommodate NLP tasks for the Indonesian Language, which is IndoBERT \cite{b27}, which has different base models ranging from 11.7M parameter to 335.2M parameter, the main model that we are going to use for doing emotion classification on product reviews.

\subsection{Comparisons}

Models based on rule-based and dictionary-based methods, as well as traditional machine learning algorithms, have been useful for emotion classification. However, BERT-based models significantly outperform them, achieving an average accuracy of 67.6\%, compared to 49.3\% for dictionary-based approaches \cite{b29}. Among BERT variants, RoBERTa demonstrated the highest recognition accuracy \cite{b30} when evaluated on the ISEAR dataset \cite{b31}.

Several studies have explored emotion classification in the Indonesian language, showing promising results. Unfortunately, its performance still falls short compared to English emotion classification \cite{b32}. Compared fine-tuned IndoBERT with traditional machine learning methods and CNNs on an Indonesian dataset, demonstrating that BERT-based models outperformed other approaches. IndoBERT achieved the highest accuracy at 79\%, highlighting its effectiveness in emotion classification for the Indonesian language. In the context of the PRDECT-ID Dataset, few studies have utilized the dataset, details can be seen in Table 2, Accuracy of 73\% was achieved using IndoBERT large-p1, while a weighted stacked generalization between Bi-LSTM and IndoBERT base-p2 reached 75\% \cite{b35}. Other variants has also been experimented by using IndoBERT and extending it using CNN layers which achieved 69.26\% accuracy and 65.49\% F1 score \cite{b36}. Sentiment classification using the dataset has also been done using GRU, Bi-GRU, and LSTM which achieved 93\% accuracy \cite{b37}.

\begin{table}[htbp]
\centering
\renewcommand{\arraystretch}{0.9}
\small  % or \scriptsize for even smaller font
\caption{Indonesia Emotion Classification with PRDECT ID Dataset}
\begin{tabular}{|c|c|c|}
\hline
\textbf{Method} & \textbf{Task} & \textbf{Accuracy} \\
\hline
IndoBERT Large \cite{b35} & ER & 73\% \\
Stacked Bi-LSTM \& IndoBERT \cite{b35} & ER & 75\% \\
IndoBERT + CNN Layer \cite{b36} & ER & 69.2\% \\
Bi-GRU \cite{b37} & SC & 93\% \\
\hline
\end{tabular}
\caption*{Note: ER refers to Emotion Recognition, while SC refers to Sentiment Classification.}
\label{tab1}
\end{table}

\section{Methodology}

This study leverages pre-trained models from the HuggingFace API, specifically IndoBERT and DistilBERT, which have been fine-tuned for Indonesian Emotion Classification. In addition, we apply bagging to combine IndoBERT and DistilBERT. The complete flow of our methodology is illustrated in Figures 2 and 3. Our main objective is to determine the most optimal data state for training, and additionally, to identify the best hyperparameters for each model in performing emotion classification.

\subsection{Data Collection}

Data that is used for this experiment is PRDECT-ID dataset \cite{b9}, the dataset label is validated and labeled by a clinical psychology expert \cite{b35}, which contains 5400 data, with 5 emotion labels which are Happy, Anger, Sadness, Love, and Fear, based on the Shaver's Emotion Theory \cite{b11}. The distribution of each label were imbalanced based on Figure 1, making it hard to train a model, we propose two methods that are used for making the dataset balance by doing augmentation.

\begin{figure}[htbp]
    \centering
    \includegraphics[width=0.3\textwidth]{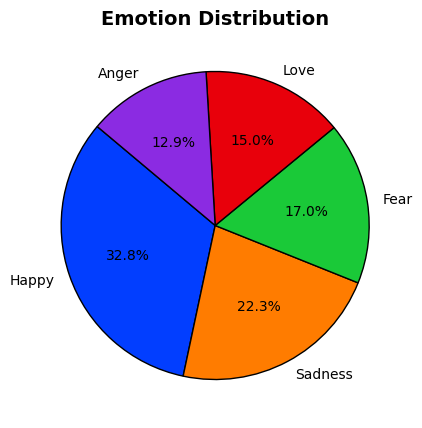} % Adjust width as needed
    \caption{Dataset Label Distribution}
    \label{fig}
\end{figure}

\subsection{Data Cleaning \& Processing}
To prepare the text data for modeling, we applied a pre-processing pipeline to remove noise and retain only meaningful information. First, we remove stop words, which are words such as "ada" (meaning "exists" or "is present"), "ia" (a neutral pronoun for "he," "she," or "it"), and "dia" (another pronoun for "he" or "she") that do not contribute significant meaning to the text. Eliminating stop words helps reduce the useless information in the dataset and improves computational efficiency because of the shorter sentence length. Next, we apply alphabet filtering, where we remove all non-alphabetic characters, including numbers, punctuation marks, and special symbols. This step prevents unnecessary noise from affecting the model's learning process. Additionally, all text is converted to lowercase to maintain consistency and prevent duplicate representations of words with different capitalizations. These preprocessing steps collectively enhance the quality of the text data by preserving only the essential linguistic information needed for effective model training.

\subsection{Data Balancing}
As told from the previous section, distribution of the dataset is not balance, which is not optimal for doing training to a model, hence we did undersampling and augmentation. In undersampling, we reduce the number of samples in each label to match the label with the fewest data points, ensuring all labels have an equal amount of data. Data augmentation is done by two method which is Back Translation and Synonym Replacement which helps improve the model performance when the dataset is not balance. Back Translation (BT) was done by translating the text in Indonesia to English and then back to Indonesia, additionally BT was also done to Arabic, but still making sure the augmented result is different. Synonym Replacement, as the name implies, involves replacing words in a sentence with their synonyms. Data Augmentation resulted in a balance dataset that each label has 1,770 text following the size of the label from Happy which originally already has 1,770 text.

\subsection{Feature Extraction}
After augmenting and processing the dataset, the text is prepared for training. Since the models we use do not accept raw text, we must convert the text into a numerical format suitable for input. This feature extraction process is performed using a pre-trained tokenizer from the IndoBERT and DistilBERT models, which transforms the text into tokenized representations that the model can process effectively.

\subsection{Data Splitting}
The dataset is split into 80\% training, 10\% validation, and 10\% test to ensure effective model learning and evaluation. The training set is used for learning patterns, the validation set for tuning hyperparameters and preventing overfitting, and the test set for assessing final performance on unseen data. Stratified sampling is applied in classification tasks to maintain class balance, ensuring a fair evaluation of the model.

\begin{figure*}[htbp]
\centerline{\includegraphics[width=0.8\textwidth]{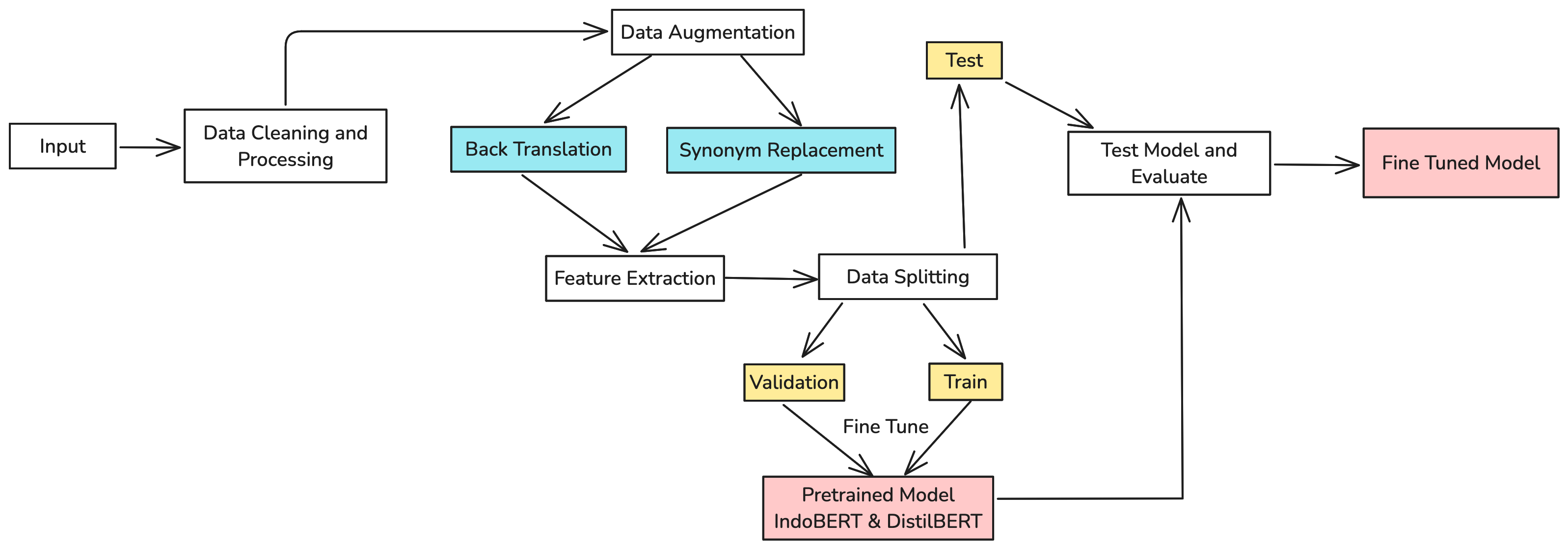}}
\caption{Methodology Flow}
\label{fig}
\end{figure*}

\subsection{Model Selection}
IndoBERT and DistilBERT were selected for the emotion classification task based on recent studies on Indonesian emotion classification, which have successfully utilized IndoBERT as a baseline model \cite{b35} \cite{b36}. DistilBERT, our second model, is a lightweight variant of BERT that has been fine-tuned for the Indonesian language. In this study, we aim to evaluate whether DistilBERT maintains competitive performance while offering a more efficient and computationally effective alternative for training.

\subsection{Finetuning}
Each pre-trained model, IndoBERT and DistilBERT, is obtained using the Hugging Face API. After feature extraction and data splitting, the processed dataset is fed into the pretrained model, allowing it to learn meaningful patterns. Once training is complete, the fine-tuned model is evaluated on the test dataset using accuracy, F1-score, recall, and precision to assess its performance, to view the details for the fine tuning can be seen in Figure 2.

Additionally, we conduct hyperparameter tuning to determine the optimal configuration for fine-tuning these models. Specifically, we experiment with different values for the number of epochs, batch size, weight decay, early stopping criteria, dropout probability, and learning rate. These adjustments aim to improve model accuracy while minimizing overfitting, particularly given the limited dataset size, which increases the risk of overfitting. Before hyperparameter tuning, the model is trained using a default configuration. The details of this configuration include a batch size of 8, a learning rate of 0.000002, and no dropout or weight decay.

\subsection{Hyperparameter Tuning}
After determining the optimal dataset configuration for IndoBERT and DistilBERT, we performed hyperparameter tuning to identify the best training settings for achieving optimal performance in each model. The tuned hyperparameters include batch size, learning rate, weight decay, and dropout probability.

\subsection{Bagging}
To enhance model performance, we applied ensemble learning using the bagging technique to combine IndoBERT and DistilBERT. Bagging was chosen to mitigate overfitting, as each individual model tended to overfit the data. By employing bagging, we aimed to reduce model bias and improve generalization.

In this approach, we selected \textit{N} models, each trained on a different resampled subset of the training dataset, details of bagging can be seen on Figure 3. Initially, we experimented with ensembles of two and three models. Later, we extended our experiments to an ensemble of ten models with varying architectures. For each model in the ensemble, we used the best-performing hyperparameters identified during previous tuning.

\begin{figure*}[htbp]
\centerline{\includegraphics[width=0.6\textwidth]{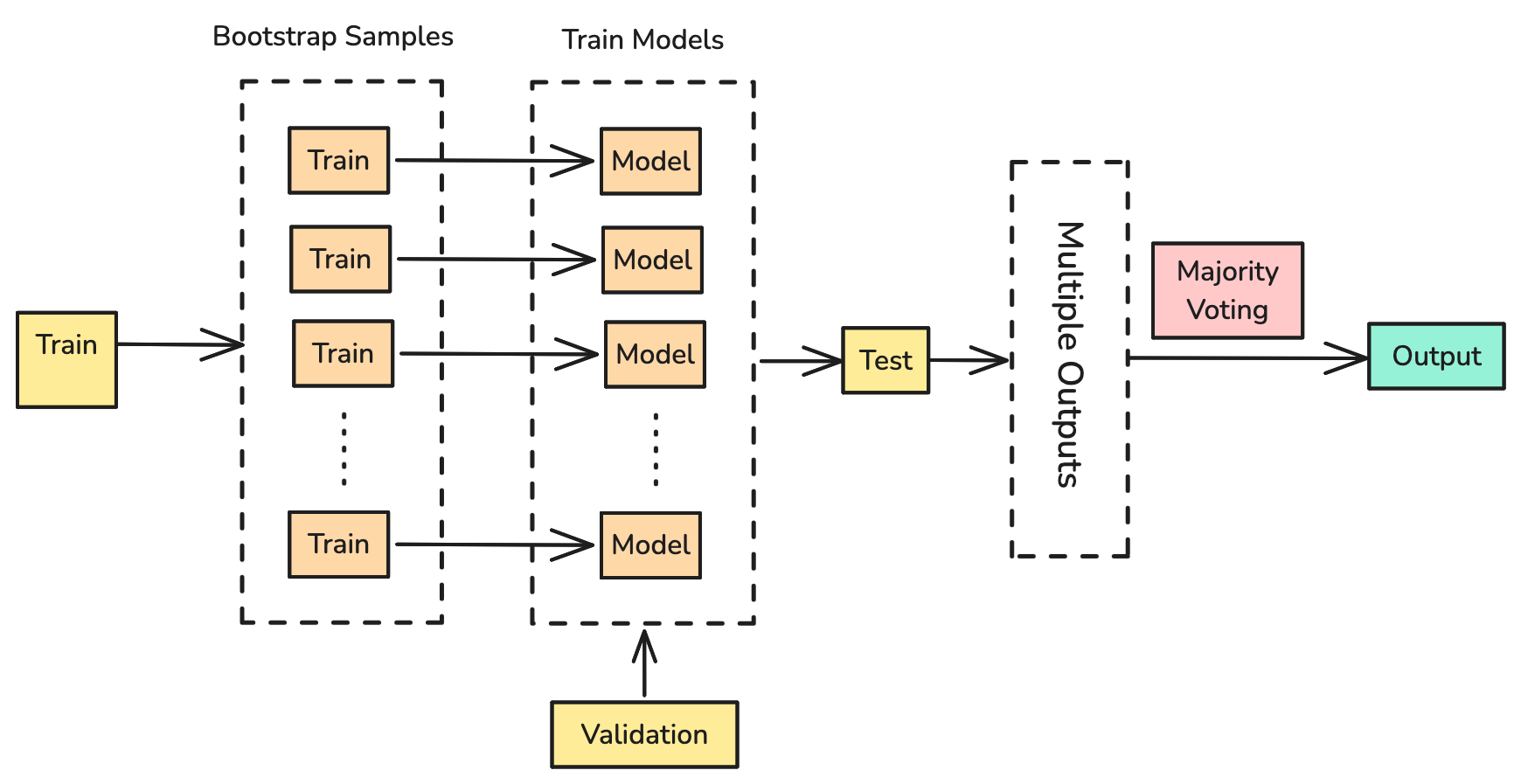}}
\caption{Bagging Flow Fine Tuning}
\label{fig}
\end{figure*}

\section{Results}
In this chapter, we present the results of our experiments on emotion classification using IndoBERT and DistilBERT. We analyze the model performance based on key evaluation metrics, compare different configurations, and discuss the learning of the ensemble impact. Additionally, we provide an error analysis to highlight potential challenges and areas for improvement.

\subsection{Baseline Model Performance}
Based on the experimental setup described in Chapter 3, our initial experiment was conducted without data augmentation or stop word removal. Initially, we compared BERT and IndoBERT to assess the performance gains of using a model specifically trained on an Indonesian dataset. The results indicate that IndoBERT, which reaches a performance of 67.78\% accuracy that outperforms BERT, which achieved 63.33\% accuracy, though the improvement is relatively modest, around 2-4\%. Furthermore, our experiments with undersampling (US) demonstrate that a balanced dataset leads to better learning, as the model achieved similar accuracy despite the reduced dataset size. The detailed results of these experiments are presented in Table 3.

%BASE MODEL RESULTS
\begin{table*}[htbp]
\centering
\small % Reduce font size
\setlength{\tabcolsep}{2pt} % Reduce space between columns
\renewcommand{\arraystretch}{1.1} % Reduce row height

\begin{minipage}{0.48\textwidth}
    \centering
    \caption{Emotion Classification without Augmentation}
    \begin{tabular}{|c|c|c|c|c|c|c|}
    \hline
    \textbf{Method} & \textbf{Acc.} & \textbf{Prec.} & \textbf{Rec.} & \textbf{F1} & \textbf{Ep.} & \textbf{US} \\
    \hline
    BERT & 0.6333 & 0.5894 & 0.5797 & 0.5812 & 3 & No \\
    IndoBERT & 0.6537 & 0.6106 & 0.6111 & 0.6056 & 3 & No \\
    IndoBERT & 0.6778 & 0.6778 & 0.6389 & 0.6338 & 4 & No \\
    IndoBERT & 0.6685 & 0.6729 & 0.6685 & 0.6701 & 4 & Yes \\
    IndoBERT & 0.6828 & 0.6891 & 0.6828 & 0.6825 & 8 & Yes \\
    \hline
    \end{tabular}
    \label{tab1}
\end{minipage}
\hfill
\begin{minipage}{0.48\textwidth}
    \centering
    \caption{Emotion Classification After Augmentation}
    \begin{tabular}{|c|c|c|c|c|c|c|}
    \hline
    \textbf{Model} & \textbf{Epoch} & \textbf{Loss} & \textbf{Acc.} & \textbf{Prec.} & \textbf{Rec.} & \textbf{F1} \\
    \hline
    IndoBERT & 4  & 1.1482  & 0.7751  & 0.7774  & 0.7751  & 0.7756  \\
    IndoBERT & 10  & 1.6305  & 0.7876  & 0.7885  & 0.7876  & 0.7876  \\
    IndoBERT & 20  & 2.0325  & 0.7740  & 0.7751  & 0.7740  & 0.7744  \\
    DistilBERT & 4  & 1.0811  & 0.7548  & 0.7564  & 0.7548  & 0.7551  \\
    DistilBERT & 10  & 1.7057  & 0.7638  & 0.7682  & 0.7638  & 0.7643   \\
    \hline
    \end{tabular}
    \label{tab:emotion_classification}
\end{minipage}
\end{table*}

% PICTURES
\begin{figure}[htbp]
    \centering
    \begin{subfigure}[b]{0.48\textwidth}
        \centering
        \includegraphics[width=\textwidth]{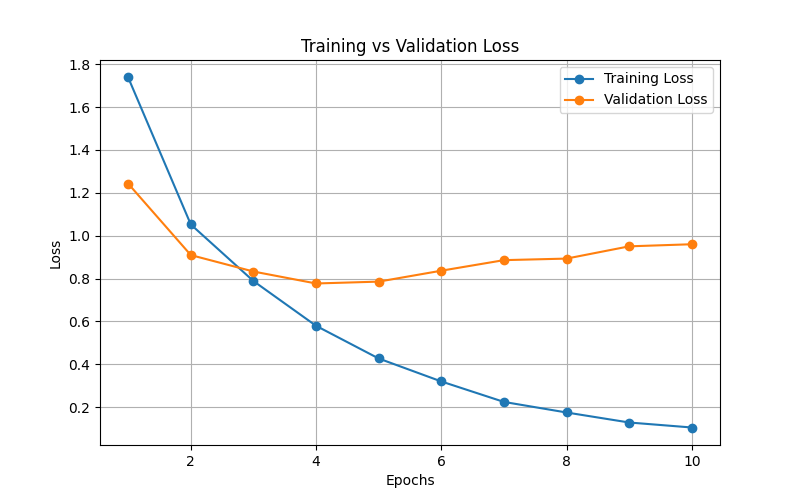}
        \caption{DistilBERT default configuration 10 Epochs}
    \end{subfigure}
    \hfill
    \begin{subfigure}[b]{0.48\textwidth}
        \centering
        \includegraphics[width=\textwidth]{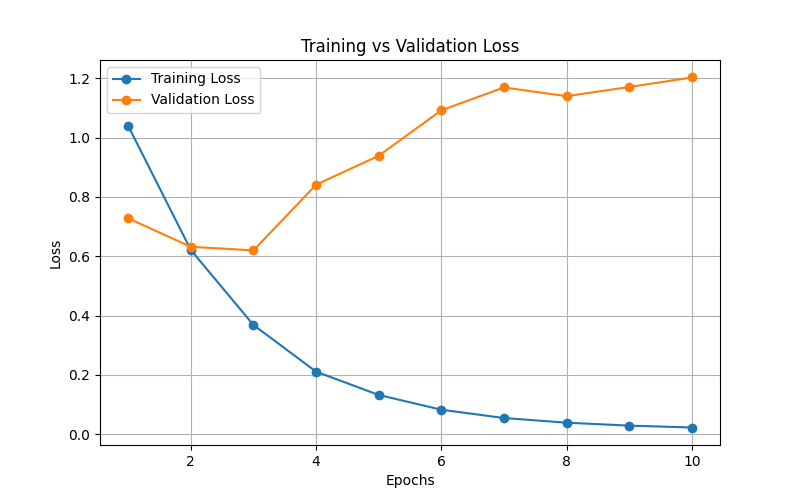}
        \caption{IndoBERT default configuration 10 Epochs}
    \end{subfigure}
    \caption{Comparison of DistilBERT and IndoBERT default configurations over 10 epochs}
    \label{fig:bert_comparison}
\end{figure}

After doing augmentation, we started introducing DistilBERT to the dataset that has been augmented. Based on the result, augmentation does improve the model performance by around 10\%, where at the highest, IndoBERT achieved 78.76\% accuracy, and DistilBERT achieved 76.38\% accuracy. Other performance in the training setting can be seen in Table 4.

\subsection{Hyperparameter Tuning}

Hyperparameter tuning involves experimenting with different training configurations to mitigate overfitting and improve model performance. In this study, tuning was conducted under two dataset settings: one with default augmentation and another with stopwords removed. The results indicate that both IndoBERT and DistilBERT perform best when stopwords are retained. Removing stopwords appears to degrade the contextual integrity of sentences, ultimately impairing the model’s ability to understand and classify them effectively.

Among the tested hyperparameters on dataset without removing stop words, details can be seen in Table 5, weight decay and epoch count had the most significant impact on performance. In the case of IndoBERT, tuning these parameters enabled the model to achieve 80\% accuracy, the highest across all configurations. This demonstrates that weight decay effectively enhances model generalization in this context. The number of training epochs significantly impacts model performance, with accuracy generally improving as epochs increase. However, a notable trade-off is observed: while accuracy rises, evaluation loss also increases. This trend is expected, as shown in Figures 4 and 5, where validation loss continues to climb with more training epochs. This suggests that while the model becomes more confident in its predictions, it may also be overfitting to the training data, leading to diminished generalization performance.

Batch size adjustments had minimal impact on performance. IndoBERT remained largely unaffected, while DistilBERT’s accuracy dropped by 4\%, indicating greater sensitivity to batch size variations. Similarly, changing the dropout probability had little to no effect on model performance. DistilBERT showed only slight accuracy improvements, while IndoBERT’s variations were negligible.

\begin{table*}[htbp]
\centering
\renewcommand{\arraystretch}{0.9}
\small  % Use \scriptsize for even more compact size
\caption{Hyperparameter Tuning Results Without Stop Words Removal}
\begin{tabular}{|c|c|c|c|c|c|c|}
\hline
\textbf{Model} & \textbf{Epochs} & \textbf{Dropout} & \textbf{Weight Decay} & \textbf{Batch Size} & \textbf{Eval Loss} & \textbf{Eval Accuracy} \\
\hline
IndoBERT & 5  & 0.1 & 0.3  & 8  & 0.9633  & 0.7876  \\
IndoBERT & 5  & 0.3 & 0.01 & 8  & 0.9695  & 0.7864  \\
IndoBERT & 5  & 0.5 & 0.01 & 8  & 0.9695  & 0.7864  \\
IndoBERT & 5  & 0.1 & 0.01 & 8  & 0.9590  & 0.7831  \\
IndoBERT & 5  & 0.1 & 0.01 & 16 & 0.7491  & 0.7819  \\
IndoBERT & 5  & 0.3 & 0.01 & 32 & 0.6486  & 0.7695  \\
IndoBERT & 10 & 0.1 & 0.01 & 16 & 1.1265  & 0.7966  \\
IndoBERT & 10 & 0.5 & 0.01 & 8  & 1.2965  & 0.7932  \\
IndoBERT & 10 & 0.3 & 0.01 & 32 & 0.9782  & 0.7887  \\
IndoBERT & 10 & 0.1 & 0.01 & 8  & 1.3320  & 0.7864  \\
IndoBERT & 10 & 0.3 & 0.01 & 8  & 1.3446  & 0.7864  \\
IndoBERT & 10 & 0.1 & 0.3  & 8  & 1.2667  & 0.8000  \\
DistilBERT & 5  & 0.1 & 0.01 & 8  & 0.7998  & 0.7345  \\
DistilBERT & 5  & 0.3 & 0.01 & 8  & 0.8018  & 0.7333  \\
DistilBERT & 5  & 0.1 & 0.3  & 8  & 0.8001  & 0.7299  \\
DistilBERT & 5  & 0.1 & 0.01 & 16 & 0.7678  & 0.7028  \\
DistilBERT & 5  & 0.3 & 0.01 & 32 & 0.8299  & 0.6734  \\
DistilBERT & 5  & 0.5 & 0.01 & 8  & 0.7990  & 0.7356  \\
DistilBERT & 10 & 0.5 & 0.01 & 8  & 1.3307  & 0.7469  \\
DistilBERT & 10 & 0.1 & 0.3  & 8  & 1.3217  & 0.7435  \\
DistilBERT & 10 & 0.1 & 0.01 & 16 & 0.9021  & 0.7412  \\
DistilBERT & 10 & 0.3 & 0.01 & 8  & 1.3187  & 0.7514  \\
DistilBERT & 10 & 0.3 & 0.01 & 32 & 0.8321  & 0.7119  \\
\hline
\end{tabular}
\label{tab:hyperparam_tuning}
\end{table*}

Efforts to improve hyperparameter tuning performance was added by adding text processing by removing stop words from the dataset, details can be seen in Table 6. Results show a really big loss of performance from both model, our hypothesis is that removing stop words from the dataset remove important context from texts making the text lose many context from the sentence.

\begin{table}[htbp]
\centering
\renewcommand{\arraystretch}{1}
\setlength{\tabcolsep}{2pt}
\caption{IndoBERT Hyperparameter Tuning After Stop Words Removal}
\begin{tabular}{|c|c|c|c|c|c|}
\hline
\textbf{Epochs} & \textbf{Dropout} & \textbf{Weight Decay} & \textbf{Batch Size} & \textbf{Eval Loss} & \textbf{Eval Accuracy} \\
\hline
4  & 0.1 & 0.01 & 16 & 0.7677 & 0.7164 \\
4  & 0.1 & 0.01 & 8  & 0.8821 & 0.7243 \\
4  & 0.1 & 0.3  & 8  & 0.9028 & 0.7198 \\
4  & 0.3 & 0.01 & 8  & 0.9041 & 0.7175 \\
4  & 0.5 & 0.01 & 8  & 0.9041 & 0.7175 \\
10 & 0.1 & 0.01 & 16 & 1.2566 & 0.7401 \\
10 & 0.1 & 0.01 & 8  & 1.4825 & 0.7356 \\
10 & 0.1 & 0.3  & 8  & 1.5355 & 0.7266 \\
10 & 0.3 & 0.01 & 8  & 1.5154 & 0.7311 \\
10 & 0.5 & 0.01 & 8  & 1.5155 & 0.7311 \\
\hline
\end{tabular}
\label{tab:hyperparam_tuning}
\end{table}

\begin{table}[htbp]
\centering
\caption{Performance of Bagging}
\begin{tabular}{|c|c|c|c|}
\hline
\textbf{Model Combination} & \textbf{Accuracy} & \textbf{F1-Score} & \textbf{Epoch} \\
\hline
IndoBERT + DistilBERT & 0.7718 & 0.7722 & 5 \\
Two IndoBERT & 0.7729 & 0.7737 & 5 \\
5 IndoBERT & 0.7887 & 0.7891 & 8 \\
5 IndoBERT & 0.7977 & 0.7979 & 10 \\
\hline
\end{tabular}
\label{tab:indoBERT_performance}
\end{table}

\subsection{Bagging}

The results from hyperparameter tuning were used to determine the best training configurations for each model, which were then combined using bagging. The findings indicate that increasing the number of IndoBERT models and extending the training epochs lead to improved performance, with the highest accuracy reaching 0.7977. However, bagging did not yield significant improvements when combining IndoBERT with DistilBERT. This outcome aligns with the hyperparameter tuning results, where IndoBERT consistently outperformed DistilBERT, suggesting that IndoBERT is inherently better suited for this task. The details can be seen in Table 7.

\section{Conclusions}
Our study aims to leverage text-based Indonesia emotion classification using PRDECT-ID dataset by integrating a new method in data processing and augmentation, evaluating each pretrained model IndoBERT and DistilBERT individually and in ensemble configuration using bagging.

Experiment results show IndoBERT superiority when being compared to DistilBERT that has been fine-tuned for emotion classification for Indonesian language, additionally IndoBERT achieves the best accuracy overall by using augmented dataset that still has stop words intact; this shows the importance of context from each sentence and the impact of augmentation for data balancing is shown by 10\% accuracy jump before augmentation. By doing hyperparameter tuning we experiment with different training configuration by achieved at the highest 80\% accuracy using IndoBERT.

Attempt to improve model performance is done by bagging where it's not successful to achieve higher performance getting the highest accuracy at 79.77\% accuracy by combining 5 IndoBERT, which is really close to the highest result we got. Future works should focus on fixing the overfitting issues, experiment with different model that can excel more on lower size of dataset while not have high overfitting. Additionally using different combination of model and configuration to get better results.

\section*{Author Contribution}
William Christian: Conduct experiment, data processing, review and editing; Daniel Adamlu: Conduct experiment, review and editing; Adrian Yu: Data gathering, review and editing. Derwin Suhartono: supervision;

\section*{Acknowledgement}
The authors would like to express their sincere gratitude for providing the facilities and support to Bina Nusantara University research grant numbered 069B/VRRTT/III/2024.

\section*{Data Availability}

The dataset used in this study is the PRDECT Dataset, which is publicly available at the following GitHub repository: \url{https://github.com/rhiosutoyo/PRDECT-ID-Indonesian-Product-Reviews-Dataset}. For further details regarding the experiments or access to the source code, please contact William Christian via email.


%% References

%The citation must be used in following style: \cite{article-minimal} \cite{article-full} \cite{article-crossref} \cite{whole-journal}.

%\bibliography{xampl}

\begin{thebibliography}{00}

\bibitem{b1} S. Peng, L. Cao, Y. Zhou, Z. Ouyang, A. Yang, X. Li, W. Jia, and S. Yu, "A survey on deep learning for textual emotion analysis in social networks," Digit. Commun. Netw., vol. 8, no. 5, pp. 745–762, 2022, doi: 10.1016/j.dcan.2021.10.003.
\bibitem{b2} S. K. Bharti, B. Vachha, R. K. Pradhan, K. S. Babu, and S. K. Jena, "Sarcastic sentiment detection in tweets streamed in real time: A big data approach," Digit. Commun. Netw., vol. 2, no. 3, pp. 108–121, 2016, doi: 10.1016/j.dcan.2016.06.002.
\bibitem{b3} Islam, M.M., Nooruddin, S., Karray, F. and Muhammad, G., 2024. Enhanced multimodal emotion recognition in healthcare analytics: A deep learning based model-level fusion approach. Biomedical Signal Processing and Control, 94, p.106241.
\bibitem{b4} Kaur, G. and Sharma, A., 2023. A deep learning-based model using hybrid feature extraction approach for consumer sentiment analysis. Journal of big data, 10(1), p.5.
\bibitem{b5} Maier, U. and Klotz, C., 2022. Personalized feedback in digital learning environments: Classification framework and literature review. Computers and Education: Artificial Intelligence, 3, p.100080.
\bibitem{b6} Alhuzali, H. and Ananiadou, S., 2021. SpanEmo: Casting multi-label emotion classification as span-prediction. arXiv preprint arXiv:2101.10038.
\bibitem{b7} Deng, J., \& Ren, F. (2021). A survey of textual emotion recognition and its challenges. IEEE Transactions on Affective Computing, 14(1), 49-67.
\bibitem{b33} Sutoyo, R., Achmad, S., Chowanda, A., Andangsari, E.W. and Isa, S.M., 2022. PRDECT-ID: Indonesian product reviews dataset for emotions classification tasks. Data in Brief, 44, p.108554.
\bibitem{b8} Warganegara, D.L. and Babolian Hendijani, R., 2022. Factors that drive actual purchasing of groceries through e-commerce platforms during COVID-19 in Indonesia. Sustainability, 14(6), p.3235.
\bibitem{b34} Hidayansyah, R., Ahmad, A. and Nabila, N.I., 2023. The impact of consumer reviews and ratings on purchase decisions on the tokopedia marketplace in indonesia. International Journal of Economics, Business, and Entrepreneurship, 6(2), pp.140-152.
\bibitem{b9} Shaver, P., Schwartz, J., Kirson, D., \& O'connor, C. (1987). Emotion knowledge: further exploration of a prototype approach. Journal of personality and social psychology, 52(6), 1061.
\bibitem{b10} Hatzivassiloglou, V. and McKeown, K., 1997, July. Predicting the semantic orientation of adjectives. In 35th annual meeting of the association for computational linguistics and 8th conference of the european chapter of the association for computational linguistics (pp. 174-181).
\bibitem{b11} Kamps, J., Marx, M., Mokken, R.J. and De Rijke, M., 2004, May. Using WordNet to measure semantic orientations of adjectives. In Lrec (Vol. 4, pp. 1115-1118).
\bibitem{b12} Mohammad, S.M. and Kiritchenko, S., 2015. Using hashtags to capture fine emotion categories from tweets. Computational Intelligence, 31(2), pp.301-326.
\bibitem{b13} Mohammad, S.M., 2017. Word affect intensities. arXiv preprint arXiv:1704.08798.
\bibitem{b14} Lewis, D.D., 1992. Feature selection and feature extraction for text categorization. In Speech and Natural Language: Proceedings of a Workshop Held at Harriman, New York, February 23-26, 1992.
\bibitem{b15} Li, P., Xu, W., Ma, C., Sun, J. and Yan, Y., 2015, June. IOA: Improving SVM based sentiment classification through post processing. In Proceedings of the 9th international workshop on semantic evaluation (SemEval 2015) (pp. 545-550).
\bibitem{b16} Talbot, R., Acheampong, C. and Wicentowski, R., 2015, June. Swash: A naive bayes classifier for tweet sentiment identification. In Proceedings of the 9th international workshop on semantic evaluation (SemEval 2015) (pp. 626-630).
\bibitem{b17} Tripathi, S., Acharya, S., Sharma, R., Mittal, S. and Bhattacharya, S., 2017, February. Using deep and convolutional neural networks for accurate emotion classification on DEAP data. In Proceedings of the AAAI Conference on Artificial Intelligence (Vol. 31, No. 2, pp. 4746-4752).
\bibitem{b18} Majumder, N., Poria, S., Hazarika, D., Mihalcea, R., Gelbukh, A. and Cambria, E., 2019, July. Dialoguernn: An attentive rnn for emotion detection in conversations. In Proceedings of the AAAI conference on artificial intelligence (Vol. 33, No. 01, pp. 6818-6825).
\bibitem{b19} Noh, S.H., 2021. Analysis of gradient vanishing of RNNs and performance comparison. Information, 12(11), p.442.
\bibitem{b20} Ge, S., Qi, T., Wu, C. and Huang, Y., 2019, June. Thu\_ngn at semeval-2019 task 3: dialog emotion classification using attentional lstm-cnn. In Proceedings of the 13th international workshop on semantic evaluation (pp. 340-344).
\bibitem{b21} Chen, J.X., Jiang, D.M. and Zhang, Y.N., 2019. A hierarchical bidirectional GRU model with attention for EEG-based emotion classification. Ieee Access, 7, pp.118530-118540.
\bibitem{b22} Zhao, J., Huang, F., Lv, J., Duan, Y., Qin, Z., Li, G. and Tian, G., 2020, November. Do RNN and LSTM have long memory?. In International Conference on Machine Learning (pp. 11365-11375). PMLR.
\bibitem{b23} Vaswani, A., Shazeer, N., Parmar, N., Uszkoreit, J., Jones, L., Gomez, A.N., Kaiser, Ł. and Polosukhin, I., 2017. Attention is all you need. Advances in neural information processing systems, 30.
\bibitem{b24} Devlin, J., Chang, M.W., Lee, K. and Toutanova, K., 2019, June. Bert: Pre-training of deep bidirectional transformers for language understanding. In Proceedings of the 2019 conference of the North American chapter of the association for computational linguistics: human language technologies, volume 1 (long and short papers) (pp. 4171-4186).
\bibitem{b25} Liu, Y., Ott, M., Goyal, N., Du, J., Joshi, M., Chen, D., Levy, O., Lewis, M., Zettlemoyer, L. and Stoyanov, V., 2019. Roberta: A robustly optimized bert pretraining approach. arXiv preprint arXiv:1907.11692.
\bibitem{b26} Sanh, V., Debut, L., Chaumond, J. and Wolf, T., 2019. DistilBERT, a distilled version of BERT: smaller, faster, cheaper and lighter. arXiv preprint arXiv:1910.01108.
\bibitem{b27} Wilie, B., Vincentio, K., Winata, G.I., Cahyawijaya, S., Li, X., Lim, Z.Y., Soleman, S., Mahendra, R., Fung, P., Bahar, S. and Purwarianti, A., 2020. IndoNLU: Benchmark and resources for evaluating Indonesian natural language understanding. arXiv preprint arXiv:2009.05387.
\bibitem{b28} Barnes, J., 2023, July. Sentiment and emotion classification in low-resource settings. In Proceedings of the 13th Workshop on Computational Approaches to Subjectivity, Sentiment, \& Social Media Analysis (pp. 290-304).
\bibitem{b29} Adoma, A.F., Henry, N.M. and Chen, W., 2020, December. Comparative analyses of bert, roberta, distilbert, and xlnet for text-based emotion recognition. In 2020 17th international computer conference on wavelet active media technology and information processing (ICCWAMTIP) (pp. 117-121). IEEE.
\bibitem{b30} Scherer, K.R. and Wallbott, H.G., 1994. Evidence for universality and cultural variation of differential emotion response patterning. Journal of personality and social psychology, 66(2), p.310.
\bibitem{b31} Setiawan, E.I., Kristianto, L., Hermawan, A.T., Santoso, J., Fujisawa, K. and Purnomo, M.H., 2021, April. Social media emotion analysis in indonesian using fine-tuning bert model. In 2021 3rd East Indonesia Conference on Computer and Information Technology (EIConCIT) (pp. 334-337). IEEE.
\bibitem{b32} Rizky, A.S. and Hidayat, E.Y., 2025. Emotion Classification in Indonesian Text Using IndoBERT. Computer Engineering and Applications Journal, 14(1), pp.1-11.
\bibitem{b35} Pramana, R., Jonathan, M., Yani, H.S. and Sutoyo, R., 2024. A Comparison of BiLSTM, BERT, and Ensemble Method for Emotion Recognition on Indonesian Product Reviews. Procedia Computer Science, 245, pp.399-408.
\bibitem{b36} Chowanda, A., Sutoyo, R., Achmad, S., Andangsari, E. W., Isa, S. M., \& Chen, T. K. Modeling Emotions Recognition on Indonesian Product Review by Combining BERT, CNN, and LSTM Architecture.
\bibitem{b37} Karunia, K., Putri, A.E., Fachriani, M.D. and Rois, M.H., 2024. Evaluation of the Effectiveness of Neural Network Models for Analyzing Customer Review Sentiments on Marketplace. Public Research Journal of Engineering, Data Technology and Computer Science, 2(1), pp.52-59.
\bibitem{b38} Azam, N., Ahmad, T., and Haq, N. U., 2021. Automatic emotion recognition in healthcare data using supervised machine learning. PeerJ Computer Science, 7, e751.
\bibitem{b39} Sridhar, A. S., and Nagasundaram, S., 2024. Customer Satisfaction Analysis based on Online Products by their Emotion Recognition using Meta Heuristic Machine Learning Algorithms. Journal of Computational Analysis and Applications, 33(2), 132–142.
\bibitem{b40} Sulaiman, S., Abdul Wahid, R., Mansor, M., Hanee, A., Ubaidullah, N. H., and Karnadi, 2024. The Application of Emotion Detection System (EmoD) in Online Learning. Journal of Advanced Research in Applied Sciences and Engineering Technology, 58(2), 196–210.

\end{thebibliography}
%\bibliographystyle{elsarticle-harv}



\end{document}